\documentclass[twocolumn]{article}[12]

\usepackage{PRIMEarxiv}
\usepackage[utf8]{inputenc} 
\usepackage[T1]{fontenc}    
\usepackage{hyperref}       
\usepackage{url}            
\usepackage{booktabs}       
\usepackage{amsfonts}       
\usepackage{nicefrac}       
\usepackage{microtype}      
\usepackage{lipsum}
\usepackage{fancyhdr}       
\usepackage{graphicx}       
\usepackage{dsfont}
\usepackage{bm} 
\usepackage{amsmath}
\usepackage{tikz}
\usepackage{pgfplots}
\usepackage{amssymb}
\usepackage{bbm}
\usepackage{subcaption}
\usepackage{tikz}
\usepackage{multirow}

\usepackage{amsfonts}
\usepackage{amssymb}
\usepackage{caption}
\usepackage{subcaption}
\usepackage{booktabs}
\usepackage{multirow}
\usepackage{graphicx}
\usepackage{xcolor}
\usepackage{svg}
\usepackage[thmmarks, amsmath, amsthm]{ntheorem}

\DeclareGraphicsExtensions{.png,.pdf}

\DeclareMathOperator*{\argmin}{arg\,min}

\usepackage{algorithmic}
\usepackage{algorithm}

\title{\textit{IMITATE}: Image Registration with Context for unknown time frame recovery}

\author{
Ziad Kheil$^{1,2,3}$, Lucas Robinet$^{1,3}$, Laurent Risser$^{2,4}$ Soleakhena Ken $^{1,2,3}$\\
$\null$\\
$\null^{1}$ Centre de Recherches en Canc\'erologie de Toulouse, INSERM UMR1037, Team RADOPT \\
$\null^{2}$ Université de Toulouse (F- 31062 Toulouse, France) \\
$\null^{3}$ Institut Universitaire du Cancer – Oncopole Claudius R\'egaud, 31059 Toulouse, France \\
$\null^{4}$ Institut de Math\'ematiques de Toulouse (UMR 5219), CNRS, Universit\'e de Toulouse, F-31062 Toulouse, France\\
}

\begin{document}

\twocolumn[
  \begin{@twocolumnfalse}
    \maketitle
    \begin{abstract}

In this paper, we formulate a novel image registration formalism dedicated to the estimation of unknown condition-related images, based on two or more known images and their associated conditions.
We show how to practically model this formalism by using a new conditional U-Net architecture, which fully takes into account the conditional information and does not need any fixed image. Our formalism is then applied to image moving tumors for radiotherapy treatment at different breathing amplitude using 4D-CT (3D+t) scans in thoracoabdominal regions. This driving application is particularly complex as it requires to stitch a collection of sequential 2D slices into several 3D volumes at different organ positions. Movement interpolation with standard methods then generates well known reconstruction artefacts in the assembled volumes due to irregular patient breathing, hysteresis and poor correlation of breathing signal to internal motion.
Results obtained on 4D-CT clinical data showcase artefact-free volumes achieved through real-time latencies. The code is publicly available at \url{https://github.com/Kheil-Z/IMITATE}.

\end{abstract}
  \end{@twocolumnfalse}
]


\section{Introduction}
\label{sec:intro}

\begin{figure*}
    \centering
  \includegraphics[width=0.95\textwidth]{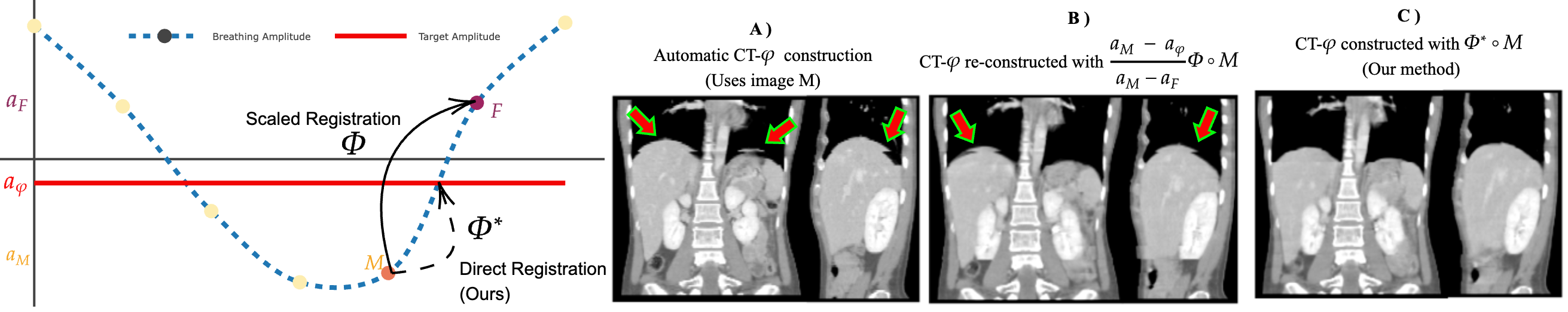}
    \caption{\textbf{(left)} Our method recovers the image motion at phase amplitudes $a_{\varphi(t)}$ using images acquired at different phases $M_i$, $i \in \{1,\cdots,n\}$. Classic approaches only use the two images, denoted $(M,F)$, acquired just before and after phase $\varphi(t)$.     \textbf{(right)} Example of recovered breathing signals for a slice at amplitude $a_{\varphi}$: (A) Default reconstructions, uses image M, (B) using $(M,F)$ with interpolated deformation, and (C) using the proposed method with $M_i$, $i \in \{1,\cdots,n\}$.}
    \label{fig:breathing_before_after}
\end{figure*}

Mobile tumors require tracking for adapted radiotherapy \cite{dsouza_use_2007}. Cine 4D-CT (3D+t) images are usually composed of $n=6$ conventional 3D CT scans at different phases of the breathing cycle to follow the movement of the patient's organs. Since no array of sensors is large enough to capture all of the organs of interest at once, frames are acquired with a moving patient table. For every table position, slices are captured over a preset duration, while a surrogate system \cite{wink_phase_2006} measures the patient's breathing amplitudes $a_{t}$ over time. The 2D images are later binned by amplitude, some are discarded, and the ones acquired at desirable amplitudes are stitched into several 3D-CTs. Due to irregular breathing \cite{watkins_patient-specific_2010}, issues arise and sum up to blatant alignment artefacts such as in Fig.~\ref{fig:breathing_before_after}-A.

In extreme cases, reconstruction artefacts can lead to patients being re-exposed to radiation for a second acquisition, with no additional guarantees of success. Furthermore, even in perfect conditions, assembling images for clinically predefined breathing phases (0\%, ... ,83\%) is ill-posed since breathing signal and images are acquired from independent systems.

Several works have attempted to tackle these shortcomings. Some focus on phase/amplitude binning \cite{abdelnour_phase_2007,yang_novel_2019}, and recently new acquisition protocols \cite{werner_intelligent_2020} have been proposed. Due to discrete sampling and irregular breathing, acquiring an image at a specific breathing amplitude remains unfeasible. Generative methods \cite{madesta_deep_2022} risk introducing non medical, artificial features. A more promising approach from the literature is to ``fix" the 4D-CT by interpolating organ movements between volumes  \cite{yang_4d-ct_2008,schreibmann_image_2006,kim_diffusion_2022,vandemeulebroucke_spatiotemporal_2011,shao_geodesic_2021,wu_reconstruction_2011}. Given two 3D images constructed around a target breathing amplitude, these approaches deformably register them, and scale the deformation field to produce a ``halfway" image. This is however limited by several strong hypotheses: organ velocity is constant and there are no differences between inhale and exhale. Furthermore, these approaches needlessly reconstruct the complete 3D image, with no possibility of only interpolating misplaced 2D slices. More importantly, they implicitly suppose the adjacent volumes are correctly constructed, although given the nature of artefacts this is often intrinsically false.
The closest approach to our work, \cite{kim_diffusion_2022} uses a diffusion paradigm to iteratively approach the required organ positions instead of scaling the deformation field obtained from a pairwise registration.
By directly aligning available images, previous methods fail to consider target specific properties.

In order to tackle these issues, we propose in our paper a novel generic and condition-based image registration formalism, and directly apply it to artefact free image recovery in 4D-CTs.
The method relies on predicting movement between images acquired at the same target location such that we can warp them into the desirable frame, using  the Deformable Image Registration (DIR) paradigm.
While this is done by using deformation fields on a moving image, as in previous works, we require no fixed image and instead use more than two moving images, each of them being associated to a condition of interest (in our application the breathing motion  amplitude).
This yields \textit{IMITATE} (IMage regIsTrAtion with conTExt), a DIR framework which
aligns moving images to a previously unseen reference using a contextual signal. The proposed implementation method uses our novel conditional U-Net architecture, and is trained by self-supervision.

\section{Methodology}
\label{sec:metho}

\noindent \textbf{Notations} Our goal is to produce anatomically correct images $\tilde{F}_{\varphi(t)} \in \mathbb{R}^{(h,w,d)}$ at the amplitude $a_{\varphi(t)}$ of the breathing motion, given a periodic signal $\varphi : t \to \varphi(t)$, where $t \in [0,T]$. Note that the breathing motion amplitude $a_{\varphi(t)}$ is acquired with an external device, as in \cite{beddar_correlation_2007}.
Specifically, let us denote $\mathcal{F}: (x,y,z,t) \to \mathcal{F}(x,y,z,\varphi(t)) \in \mathbb{R}$ the ideal images containing all voxel values over the whole support of $\varphi$ at phase $\varphi(t)$ of the breathing motion. Given $N$ measurements of $\mathcal{F}$ at specific phases $V = [\varphi_1, \cdots, \varphi_N]$, we then estimate with $\tilde{F}_{\varphi_{\epsilon}}$  the unmeasured image
$\mathcal{F}(\cdot ,\varphi_{\epsilon})$ at phase $\varphi(t)=\varphi_{\epsilon}$.\\
\noindent \textbf{Standard approach}
Let $\varphi_M$ and $\varphi_F$ $\in V$ be the phases temporally acquired just \textit{before} and \textit{after} the desired $\varphi_{\epsilon}$. For simplicity,  we note here $M,F$ the images of $\mathcal{F}$ acquired at phases $\varphi_M,\varphi_F$.  These notations are summarized Fig.~\ref{fig:breathing_before_after} (left).
The standard approach consists in first registering $M$ on $F$
\begin{align}
\Hat{\Phi} = \argmin_{\Phi}\mathcal{L}(F,M \circ \Phi) \,,
\label{eq:optim}
\end{align}
and then resampling $M$ using a \textit{linearly rescaled deformation} to phase $\varphi_\epsilon$, \textit{i.e.} $\Phi^{*} \approx \frac{\varphi_{M}-\varphi_{\epsilon}}{\varphi_{M}-\varphi_{F}} \Hat{\Phi}$ and then $\tilde{F}_{\varphi_{\epsilon}} = M \circ \Phi^{*}$.

\begin{figure*}
    \centering
  \includegraphics[width=0.9\textwidth]{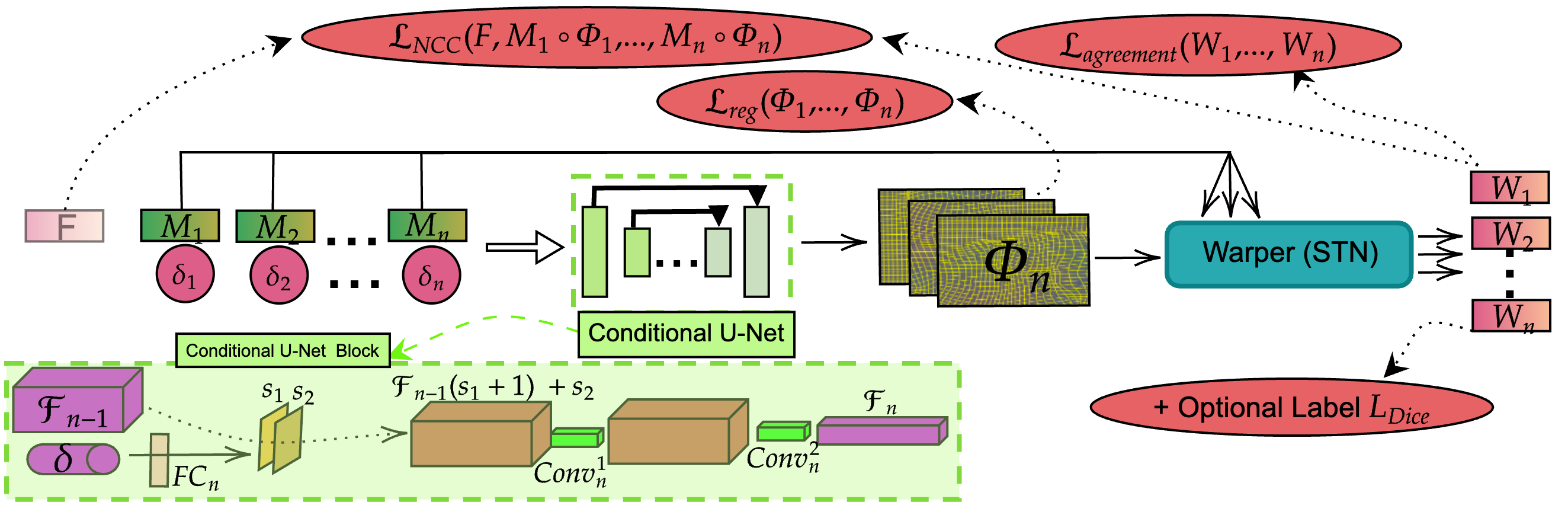}
    \caption{Main framework pipeline. F is the unseen, reference image, $M_1,...M_n$ the moving contextual frames and their associated signals $\delta_i$. Additional segmentation labels can be used like in \cite{balakrishnan_voxelmorph_2019} (Sec~\ref{sec:ref_agno_reg_frame}). Figure also displays a convolutional block from the proposed Conditional U-Net. $\mathcal{F}_{n-1}$ is the feature map at depth n, $FC_n$ the fully connected layer which produces $(s_1,s_2)$, and $Conv^{1}_n$,$Conv^{2}_n$  the convolution layers. See section \ref{sec:model}}
    \label{fig:base_framework}
\end{figure*}
The movement producing the missing frame is theoretically obtained. As shown Fig.~\ref{fig:breathing_before_after}-(B), this strategy is however known to produce reconstruction artefacts in the areas with large and fast organ motion.

\noindent \textbf{Proposed approach}
Our key contribution is to reformulate the image registration problem by computing $\Phi^{*}$ using \textit{more than two phase-related images} and \textit{conditionally on non-image features}. Our strategy, denoted \textit{IMITATE} (IMage regIsTrAtion with conTExt), relies on a conditional U-Net architecture (see Fig.~\ref{fig:base_framework}), making the problem computationally efficient.
It makes use of target amplitudes or phases as conditional variables to predict deformation fields, effectively aligning organs to their required position, \textit{i.e.} to the unseen \textit{reference}.

\subsection{Reference Agnostic Registration Framework}
\label{sec:ref_agno_reg_frame}

\noindent \textbf{Conditioning variables}
We first consider the $n \geq 2$ images $M_i$, $i \in \{1, \ldots,n\}$ acquired at times $t_i$. Corresponding phases and amplitudes are $\varphi_{i}$ and  $a_{\varphi_{i}}$, respectively. We now define $n$ conditioning variables  $\delta_i$, each of them being also associated to an image $M_i$. In our problem $\delta_i = a_{\varphi_{i}}-a_{\varphi_{\epsilon}}$, but
$\delta_i = \varphi_{M_i}-\varphi_{\epsilon}$ could also be an eligible conditioning variable. In the general case, the conditioning variables $\delta_i$, $i \in \{1, \ldots,n\}$ model a notion of proximity between acquisition properties at times $t_i$ and the corresponding property for the image to recover.
Their values also have to be continuous and differentiable with respect to acquisition times.

\noindent \textbf{Reference Agnostic motion estimation}
The IMITATE framework directly predicts the deformation field $\Phi^{*}$ by using the target signal value $\varphi_{\epsilon}$  and $n \in [1,N]$ adjacent, available images. A DNN $f_{\theta}$, with parameters $\theta$, is optimized to estimate the deformations to transport the $n$ contextual images $\bar{M}= \{M_i\}_{i=1}^{n}$ to the unobserved image $\mathcal{F}(\cdot ,\varphi_{\epsilon})$ conditionally on $\bar{\delta} = \{\delta_i\}_{i=1}^{n}$, \textit{i.e.}
$f_{\theta}(\bar{M},\bar{\delta}) = \bar{\Phi} = \{\Phi_i\}_{i=1}^{n}$.

Image recovery at condition $\varphi_{\epsilon}$ is then performed by first selecting the image $M^{*}$ among the $\{M_i\}_{i=1}^{n}$ associated to the smallest $\delta_i$. Then the corresponding deformation  $\Phi^{*}=\Phi_i$ is used to estimate $\mathcal{F}(\cdot ,\varphi_{\epsilon})$, \textit{i.e.}:

\begin{align}
\tilde{F}_{\varphi_{\epsilon}} = M^{*} \circ \Phi^{*} 
 \label{eq_phi}
\end{align}
It is important to remark here that although a single observation $M_i$ is used in  Eq. \eqref{eq_phi}, the deformation $\Phi^{*}$ is computed by considering all contextual information $(\bar{M},\bar{\delta})$. The benefit of using all of this information is developped in
Section \ref{sec:train_inference} when formalizing the training loss.

\noindent \textbf{Analogy with pairwise DIR}
In the case where $n=2$, the network still operates  following Eq. \eqref{eq_phi}, which is similar to a pairwise registration. Using $n>2$ helps account for variations in patients' breathing patterns, and introduces organ motion across different breathing amplitudes, giving the network cinematic information about the target.

\subsection{Training and Inference}
\label{sec:train_inference}

The aforementioned registration framework requires a specific training objective tailored to produce smoothly varying, and anatomically accurate slices.

\noindent \textbf{Mini-batch observations}
As usual when training neural networks, we adopt a standard gradient descent-based approach to optimize the parameters $\theta$. Each mini-batch observation is made as follows. A target $F$ is randomly selected out of all known images. For readability purpose, we use the same notations as in Section \ref{sec:ref_agno_reg_frame} and denote $\bar{M}=(M_1, ..., M_n)$ the remaining images. Moving images $(M_1, ..., M_n)$, acquired before and after $F$, are stacked and forwarded through the network, which outputs $n$ deformation fields
$f_{\theta}(\bar{M},\bar{\delta}) = \{\Phi_i\}_{i=0}^{n}$. These fields warp the input images: $W_{i} = M_{i} \circ \Phi_{i}$. Note that if segmentation masks  $S^{M_i}$ are known, they are also warped: $S^{W_i} = S^{M_i} \circ \Phi_{i} $, aiming to match the fixed target mask $S^{F}$ as done in \cite{balakrishnan_voxelmorph_2019}.

\noindent \textbf{Loss Terms} The training objective is to match the warped images to the held-out target image. We design a loss function consisting of a weighted combination of four sub-losses:
\begin{align}
\mathcal{L}(\bar{W},\bar{S^{W}},\bar{\Phi},F,S^{F}) = &\alpha \mathcal{L}_{sim}(F,\bar{W})  + \beta \mathcal{L}_{struct}(S^{F},\bar{S^{W}})\notag \\ &+ \gamma \mathcal{L}_{reg}(\bar{\Phi}) + \zeta \mathcal{L}_{agreement}(\bar{W}) \label{eq1}
\end{align}
The similarity and structural losses (Negative Cross Correlation and Dice) enforce pixel-wise and organ shape similarity between the warped and target images. Since multiple warped images are output, we average their similarity and segmentation mask losses:
$\mathcal{L}_{sim}(F,\bar{W}) = \frac{1}{n}\sum_{i=0}^{n} \mathcal{L}_{NCC}(F,W_i)$, and $\mathcal{L}_{struct}(S^{F},\bar{S^{W}}) = \frac{1}{n}\sum_{i=0}^{n} \mathcal{L}_{dice}(S^{F},S^{W_i})$.

Additionally, regularization terms are used to generate sound deformation fields, due to the ill-defined nature of image registration. The regularization of the predicted fields is averaged:
$\mathcal{L}_{reg}(\bar{\Phi}) = \frac{1}{n}\sum_{i=0}^{n} \mathcal{L}_{DetJac}(\Phi_i)$ ,
with $\mathcal{L}_{DetJac}(\Phi_i) = \mathcal{L}_{MSE}(DJ_{i},\mathbbm{1}) $ where
$DJ_{i} = \det \mathbb{J}(\Phi_i)$ is the determinant of the jacobian of the deformation field. This term, adapted from \cite{rohlfing_volume-preserving_2003}, promotes small, volume-preserving deformations without foldings.

Finally, a prediction agreement loss ensures consistency of the predicted deformation fields: $\mathcal{L}_{agreement}(\bar{W}) = \frac{1}{n}\sum_{i=0}^{n} \mathcal{L}_{MSE}(\widehat{W} - W_i) $ where $\widehat{W}(n,m) = \frac{1}{n}\sum_{i=0}^{n} W_{i}(n,m)$ is the average warped image.

\begin{table*}[ht]
\caption{Reconstruction metrics: Models are defined as (fixed image, num. moving images, $\delta$ encoding dimension). Best results are bold, second best italicized. Wilcoxon signed rank test shows IMITATE achieves $p<1e^{-5}$ for all metrics except $\ast$, where $p<0.005$. Incremental improvements highlight each component's utility: contextual frames, conditional UNet, and reference-agnostic training. }\label{tab3}
\begin{tabular}{|l|c|c|c|c|}
\hline
\textbf{Model} & \textbf{NCC}   ($\downarrow$)           & \textbf{N-NCC}   ($\downarrow$)        & \textbf{RMSE} ($\downarrow$)           & \textbf{N-RMSE}    ($\downarrow$)      \\ \hline
(1,1,-) $\rightarrow$ Pairwise                 & -0.2423 ± 0.0492          & 0.0547 ± 0.0167          & 0.0908 ± 0.0482          & 0.0430 ± 0.0314          \\ 
(1, 3, -) $\rightarrow$ Multi-Registration          & -0.3294 ± 0.0185          & 0.0401 ± 0.0091(*)         & 0.0468 ± 0.0087         & 0.0223 ± 0.0080          \\ 
\textit{(1, 3, 32)}  $\rightarrow$ \textit{Breathing Amps.}        & \textit{-0.3295 ± 0.0185} & \textit{0.0399 ± 0.0090}(*) & \textit{0.0466 ± 0.0087} & \textit{0.0222 ± 0.0080} \\ 
\textbf{(-, 3, 32)} $\rightarrow$ \textbf{IMITATE} & \textbf{-0.3399±0.0189} & \textbf{0.0375±0.0093} & \textbf{0.0427±0.0079} & \textbf{0.0181 ± 0.0071} \\ \hline
\end{tabular}
\end{table*}

\subsection{Conditional U-Net model}
\label{sec:model}
Since the introduction of Spatial Transformer Networks (STNs) \cite{jaderberg_spatial_2016}, and their integration with with VoxelMorph \cite{balakrishnan_voxelmorph_2019}, the U-Net \cite{navab_u-net_2015} architecture has become staple of deep learning based DIR. Although the fixed image is omitted, IMITATE functions similarly, with moving images stacked channel-wise. We adapt the U-Net encoder-decoder structure to handle additional signals relating to every moving image.
this is done similarly as the temporal encoding step used in latent diffusion tasks \cite{ho_denoising_2020}. The signal values associated to the images are stacked and encoded to a fixed-size representation using sinusoidal position embedding scheme introduced in \cite{vaswani_attention_2023}.  Before each convolution block,  an additional MLP layer in the block derives scale and shift tensors from the fixed amplitude encoding, which are then used to scale and shift the inputs of the convolutions, as illustrated in Fig.~\ref{fig:base_framework}.

This layer-specific conditioning allows the U-NET to predict deformation fields which fully take into account all of the input images, as well as their amplitude differences with the target at every encoder/decoder step. Note that this conditional architecture can also be used in regular registration tasks, with any U-Net backbone. In that case, the amplitude differences are those between moving images and the fixed one : $\forall i \in [0,n]$, $\delta_{i} = \varphi_{M_{i}}-\varphi_{F}$.

\subsection{Implementation Details}
The chosen backbone model was the Attention U-Net \cite{oktay_attention_2018} from the MONAI \cite{consortium_monai_2023} model library. Attention Gates allow the network to consider interactions between channels incoming through the skip connections, as they display features of organs at different positions. We implemented a conditional attention U-Net as described in Section \ref{sec:model} which was trained for registration tasks with the Eq. \eqref{eq1} objective. Using a grid search, we optimized the loss term weights for the final models: $\alpha=0.7$, $\beta=0.3$, $\gamma=0.3*0.7*0.3$ and $\zeta=0.7$ ($\alpha=0.8$,$\beta=0.8$, $\gamma=0.2*0.2$ for the classic pairwise model). The networks were optimised over 100 epochs using Adam optimizer, with a learning rate of $1e^{-3}$, and a Cosine Annealing learning rate scheduler.
Finally, in addition to the pairwise registration, we trained models with number of moving images $n$ ranging from 2 to 8. Hence tested models could use either a fixed image reference, a target amplitude, or both.

\section{Experiments and Results}
\label{sec:res}
We now apply the IMITATE framework to unknown frame recovery in CINE 4D-CT acquisitions.
Images are acquired by 2D slices along with the corresponding breathing signal. In this setting, our models are tested on 2D images.
Open-source 4D-CT datasets \cite{vandemeulebroucke_spatiotemporal_2011} provide final 3D-CTs but discard unused frames and breathing amplitudes, leaving out valuable data. To the best of our knowledge, no public 4D-CT datasets conserve all slices and breathing signals, so we trained our models on an in-house dataset.

We split 10 patients into independent training and validation sets for hyperparameter tuning, while 106 unseen patients were used for evaluation. Each patient contributed numerous training samples ($\binom{11}{n}\times 8 \times 14$ pairs of 2D slices for each).
Images were resized to $(256,256)$ and pixel intensities normalised to [0,1].
For a fair comparison of benchmarks, all methods employ an attention U-Net architecture \cite{oktay_attention_2018} tuned on the validation set.

\subsection{Supervised Results: Frame Interpolation}
IMITATE was first evaluated in a controlled, supervised setting: We artificially removed frames from the test set images, then reconstructed them using adjacent slices.More specifically, for each test set patient, at every table position, we succinctly remove a frame and reconstruct it using its neighbours.
In this case, ground-truth frames to be created are known, hence we evaluate the similarity metrics of the obtained images, dice scores of their segmentation masks, as well as the regularity of the employed deformation fields.
Table~\ref{tab2} shows a clear improvement of the interpolated images, in terms of Dice score as well as similarity metrics. Furthermore the IMITATE variant produces significantly more regular deformations, poruding smoother fields.

\begin{table}[ht]
\caption{Comparison of the proposed method with pairwise registration and scaling approach (using a trained Attention U-Net backbone.)}\label{tab2}
\centerline{
\resizebox{\columnwidth}{!}{%
\begin{tabular}{|l|c|c|c|}
\hline
Method           & NCC ($\downarrow$) & DetJac ($\downarrow$) & Dice ($\uparrow$) \\ \hline
Pairwise and Scaling  & -0.289 $\pm$ 0.023 & 0.215 $\pm$ 0.004 & 0.858 $\pm$ 0.031 \\ \hline
\textbf{3-input Imitate} & \textbf{-0.467 $\pm$ 0.052} & \textbf{0.0171 $\pm$ 0.0021} & \textbf{0.891 $\pm$ 0.030} \\ \hline
\end{tabular}
}
}
\end{table}

\subsection{Unsupervised Results: 4D-CT Artefact Correction}
We also benchmark the different methods on the final 4D-CT reconstruction task. Five target CT phases [0\%,33\%,50\%, 66\%,83\%] were reconstructed for each of the 106 test patients by different models.
All models warped the same image to produce the missing frames.
Common 4D-CT metrics from \cite{kim_study_2022,castillo_assessment_2014} are used to evaluate the spatial continuity of the volumes. These image quality metrics are reported in Table-\ref{tab3}.  IMITATE outperforms regular registration on all metrics evaluated (p-values$<$0.005 under Wilcoxon signed rank tests).

\textbf{Ablation}: Table-\ref{tab3} shows that using multiple moving images for context improves results (second line), so does using amplitude information (third line). Finally, IMITATE-style training optimizes results by targeting the correct amplitude.
\vspace{-0.3cm}
\section{Conclusion}
\vspace{-0.3cm}
We propose a novel registration approach which is better suited for frame interpolation tasks than pairwise registration. It is enabled by conditioning a U-Net on readily available, yet underutilized breathing meta-data of 4D-CT imaging. Our method explicitly targets desirable properties of a unacquired image. Experimental results display significant advantages over classic registration methods: improved reconstruction metrics and spatial regularity of deformation fields, but more importantly a  potential  cutback on the high quantity of acquisitions previously required for a 4D-CT.

Conditioning registration on additional data opens the door to enlightened registration frameworks, such as data specific to tissue elasticity, stiffness or other deformation relevant information.
\vspace{-0.3cm}
\bibliographystyle{elsarticle-num} 
\vspace{-0.2cm}
\bibliography{references}

\begin{thebibliography}{10}
\expandafter\ifx\csname url\endcsname\relax
  \def\url#1{\texttt{#1}}\fi
\expandafter\ifx\csname urlprefix\endcsname\relax\def\urlprefix{URL }\fi
\expandafter\ifx\csname href\endcsname\relax
  \def\href#1#2{#2} \def\path#1{#1}\fi

\bibitem{dsouza_use_2007}
W.~D. D’Souza, et~al., \href{https://www.sciencedirect.com/science/article/pii/S0958394707000076}{The {Use} of {Gated} and {4D} {CT} {Imaging} in {Planning} for {Stereotactic} {Body} {Radiation} {Therapy}}, Medical Dosimetry 32~(2) (2007) 92--101.
\newblock \href {https://doi.org/10.1016/j.meddos.2007.01.006} {\path{doi:10.1016/j.meddos.2007.01.006}}.
\newline\urlprefix\url{https://www.sciencedirect.com/science/article/pii/S0958394707000076}

\bibitem{wink_phase_2006}
N.~M. Wink, et~al., Phase versus amplitude sorting of {4D}-{CT} data, Journal of Applied Clinical Medical Physics 7~(1) (2006) 77--85.

\bibitem{watkins_patient-specific_2010}
W.~T. Watkins, et~al., Patient-specific motion artifacts in {4DCT}, Medical Physics 37~(6) (2010) 2855--2861.
\newblock \href {https://doi.org/10.1118/1.3432615} {\path{doi:10.1118/1.3432615}}.

\bibitem{abdelnour_phase_2007}
A.~F. Abdelnour, et~al., Phase and amplitude binning for {4D}-{CT} imaging, Physics in Medicine and Biology 52~(12) (2007) 3515--3529.

\bibitem{yang_novel_2019}
J.~Yang, et~al., A {Novel} {4D}-{CT} {Sorting} {Method} {Based} on {Combined} {Mutual} {Information} and {Edge} {Gradient}, IEEE Access 7 (2019) 138846--138856.

\bibitem{werner_intelligent_2020}
R.~Werner, et~al., Intelligent {4D} {CT} sequence scanning ({i4DCT}): {First} scanner prototype implementation and phantom measurements of automated breathing signal-guided {4D} {CT}, Medical Physics 47~(6) (2020) 2408--2412.

\bibitem{madesta_deep_2022}
F.~Madesta, et~al., Deep learning-based conditional inpainting for restoration of artifact-affected {4D} {CT} images, Medical Physics 51~(5) (2022) 3437--3454.

\bibitem{yang_4d-ct_2008}
D.~Yang, et~al., {4D}-{CT} motion estimation using deformable image registration and {5D} respiratory motion modeling, Medical Physics 35~(10) (2008) 4577--4590.

\bibitem{schreibmann_image_2006}
E.~Schreibmann, et~al., Image interpolation in {4D} {CT} using a {BSpline} deformable registration model, International Journal of Radiation Oncology*Biology*Physics 64~(5) (2006) 1537--1550.

\bibitem{kim_diffusion_2022}
B.~Kim, J.~C. Ye, Diffusion {Deformable} {Model} for {4D} {Temporal} {Medical} {Image} {Generation} (Jun. 2022).

\bibitem{vandemeulebroucke_spatiotemporal_2011}
J.~Vandemeulebroucke, et~al., Spatiotemporal motion estimation for respiratory-correlated imaging of the lungs, Medical Physics 38~(1) (2011) 166--178.

\bibitem{shao_geodesic_2021}
W.~Shao, et~al., Geodesic density regression for correcting {4DCT} pulmonary respiratory motion artifacts, Medical Image Analysis 72 (2021) 102140.

\bibitem{wu_reconstruction_2011}
G.~Wu, et~al., Reconstruction of {4D}-{CT} from a {Single} {Free}-{Breathing} {3D}-{CT} by {Spatial}-{Temporal} {Image} {Registration}, in: G.~Székely, H.~K. Hahn (Eds.), Information {Processing} in {Medical} {Imaging}, Lecture {Notes} in {Computer} {Science}, Springer, Berlin, Heidelberg, 2011, pp. 686--698.

\bibitem{beddar_correlation_2007}
A.~S. Beddar, et~al., Correlation between internal fiducial tumor motion and external marker motion for liver tumors imaged with {4D}-{CT}, International Journal of Radiation Oncology*Biology*Physics 67~(2) (2007) 630--638.

\bibitem{balakrishnan_voxelmorph_2019}
G.~Balakrishnan, et~al., {VoxelMorph}: {A} {Learning} {Framework} for {Deformable} {Medical} {Image} {Registration} (Sep. 2019).

\bibitem{rohlfing_volume-preserving_2003}
T.~Rohlfing, et~al., Volume-preserving nonrigid registration of {MR} breast images using free-form deformation with an incompressibility constraint, IEEE transactions on medical imaging 22~(6) (2003) 730--741.

\bibitem{jaderberg_spatial_2016}
M.~Jaderberg, et~al., Spatial {Transformer} {Networks} (Feb. 2016).

\bibitem{navab_u-net_2015}
O.~Ronneberger, et~al., \href{http://link.springer.com/10.1007/978-3-319-24574-4_28}{U-{Net}: {Convolutional} {Networks} for {Biomedical} {Image} {Segmentation}}, in: Medical {Image} {Computing} and {Computer}-{Assisted} {Intervention} – {MICCAI} 2015, Vol. 9351, Springer International Publishing, Cham, 2015, pp. 234--241.
\newblock \href {https://doi.org/10.1007/978-3-319-24574-4_28} {\path{doi:10.1007/978-3-319-24574-4_28}}.
\newline\urlprefix\url{http://link.springer.com/10.1007/978-3-319-24574-4_28}

\bibitem{ho_denoising_2020}
J.~Ho, A.~Jain, P.~Abbeel, Denoising {Diffusion} {Probabilistic} {Models} (Dec. 2020).

\bibitem{vaswani_attention_2023}
A.~Vaswani, et~al., Attention {Is} {All} {You} {Need}, arXiv:1706.03762 [cs] (Aug. 2023).

\bibitem{oktay_attention_2018}
O.~Oktay, et~al., Attention {U}-{Net}: {Learning} {Where} to {Look} for the {Pancreas} (May 2018).

\bibitem{consortium_monai_2023}
M.~Consortium, \href{https://zenodo.org/records/8436376}{{MONAI}: {Medical} {Open} {Network} for {AI}} (Oct. 2023).
\newblock \href {https://doi.org/10.5281/zenodo.8436376} {\path{doi:10.5281/zenodo.8436376}}.
\newline\urlprefix\url{https://zenodo.org/records/8436376}

\bibitem{kim_study_2022}
C.~Kim, et~al., A study of quantitative indicators for slice sorting in cine-mode {4DCT}, PloS One 17~(8) (2022) e0272639.

\bibitem{castillo_assessment_2014}
S.~J. Castillo, et~al., Assessment of a quantitative metric for {4D} {CT} artifact evaluation by observer consensus, Journal of Applied Clinical Medical Physics 15~(3) (2014) 190--201.

\end{thebibliography}

\end{document}